\documentclass[letterpaper]{article}
\usepackage[preprint]{aaai2027}
\usepackage[hyphens]{url}
\usepackage{graphicx}
\usepackage{natbib}
\usepackage{caption}
\usepackage{subcaption}
\usepackage{algorithm}
\usepackage{algorithmic}
\usepackage{enumitem}
\usepackage{amsmath}
\usepackage{amsfonts}
\usepackage{amssymb}
\usepackage{booktabs}
\usepackage{multirow}
\usepackage[table]{xcolor}

\newcommand{\ms}[2]{#1{\scriptsize $\pm$#2}}

\title{Reconstruction-Shift Discrimination via Mask-Guided Latent Diffusion for Medical Anomaly Detection}

\author{
    Yibo Wan\textsuperscript{\rm 1},
    Jinyu Cai\textsuperscript{\rm 2}\corresponding,
    Yunhe Zhang\textsuperscript{\rm 3},
    Yi Bin\textsuperscript{\rm 4},
    See-Kiong Ng\textsuperscript{\rm 2}
}
\affiliations{
    \textsuperscript{\rm 1}School of Computing, National University of Singapore, Singapore\\
    \textsuperscript{\rm 2}Institute of Data Science, National University of Singapore, Singapore\\
    \textsuperscript{\rm 3}Department of Computer and Information Science, SKL-IOTSC, University of Macau, China\\
    \textsuperscript{\rm 4}School of Computer Science and Technology, Tongji University, China\\
    yibo.wan19@outlook.com, \{jinyucai,seekiong\}@nus.edu.sg, zhangyhannie@gmail.com, yi.bin@hotmail.com
}

\begin{document}

\maketitle

\begin{abstract}
Unsupervised medical anomaly detection learns normal anatomical patterns from healthy training images and identifies deviations at test time. Reconstruction-based and diffusion-based methods commonly use the difference between an input image and its reconstruction as anomaly evidence. However, this residual can be ambiguous. Expressive models may preserve pathological structures, while benign anatomical variation, imaging noise, and acquisition differences may also produce large reconstruction errors. We propose discriminative mask-guided diffusion (DMD), a medical anomaly detection framework that complements residual-based localization with reconstruction-shift discrimination. DMD first learns a compact quantized latent representation of normal images. Localized masks then perturb selected latent regions, and a latent diffusion model reconstructs the perturbed representations. The resulting reconstructions are paired with their original normal images to define a self-supervised classification task. At inference, the classifier provides a learned image-level anomaly score, while the residual between the input and its diffusion-based reconstruction yields a pixel-level anomaly map. Experiments on five datasets spanning brain MRI, breast ultrasound, and chest radiography show that DMD achieves the best overall performance among the state-of-the-art baseline methods.
\end{abstract}

\section{Introduction}
\label{sec1}
Medical anomaly detection~\cite{RN1} aims to identify images that contain pathological findings and to localize the affected regions. It is especially important in brain MRI, breast ultrasound, and chest radiography, where abnormalities may be subtle, low-contrast, and highly heterogeneous~\cite{lundervold2019overview,zhang2024global,nahiduzzaman2025hybrid}. Unlike fully supervised classification or segmentation, unsupervised medical anomaly detection assumes that only normal images are available during training. This normal-only setting is practically relevant because detailed pathological annotations are costly to obtain and often limited to specific disease categories~\cite{RN2,cheng2025multi}. It also creates a fundamental challenge. The model must identify deviations using only the variability observed in normal training data.

A dominant approach is to learn a reconstruction model from normal images and use reconstruction errors as anomaly evidence. Early methods employ autoencoders~\cite{atlason2019unsupervised}, variational autoencoders~\cite{RN12}, and generative adversarial networks~\cite{RN8}. More recent methods use denoising networks and diffusion models to reconstruct inputs according to patterns learned from normal data~\cite{ho2020denoising,RN2,rombach2022high}. Latent diffusion models perform denoising in a compressed representation, which reduces the spatial resolution of the diffusion process~\cite{rombach2022high}. These methods commonly measure the residual between an input and its reconstruction. Although residuals provide valuable spatial cues, they are affected by two opposing failure modes. An expressive model may reproduce pathological structures and yield weak responses over true lesions. Normal anatomical variation, acquisition differences, and reconstruction artifacts may instead produce large responses in healthy regions. Aggregating such residuals into a single image-level score can therefore mix pathological changes with benign reconstruction errors. This motivates us to explore a learned image-level signal that complements rather than replaces residual-based localization.

Recent studies provide two useful directions for addressing this problem. Denoising and masking methods construct restoration tasks by corrupting selected regions and recovering them from the surrounding image context~\cite{kascenas2023role,liang2024itermask,liang2025itermask3d,beizaee2025mad}. In parallel, self-supervised discriminative methods transform normal images to create training signals for anomaly-sensitive classifiers~\cite{li2021cutpaste,zavrtanik2021draem,schluter2022natural}. Together, these directions motivate a direct question. \textit{\textbf{Can reconstruction shifts provide self-supervision for image-level discrimination while residuals retain spatial cues for localization?}}

\begin{figure*}[ht]
\centering
\includegraphics[width=\linewidth]{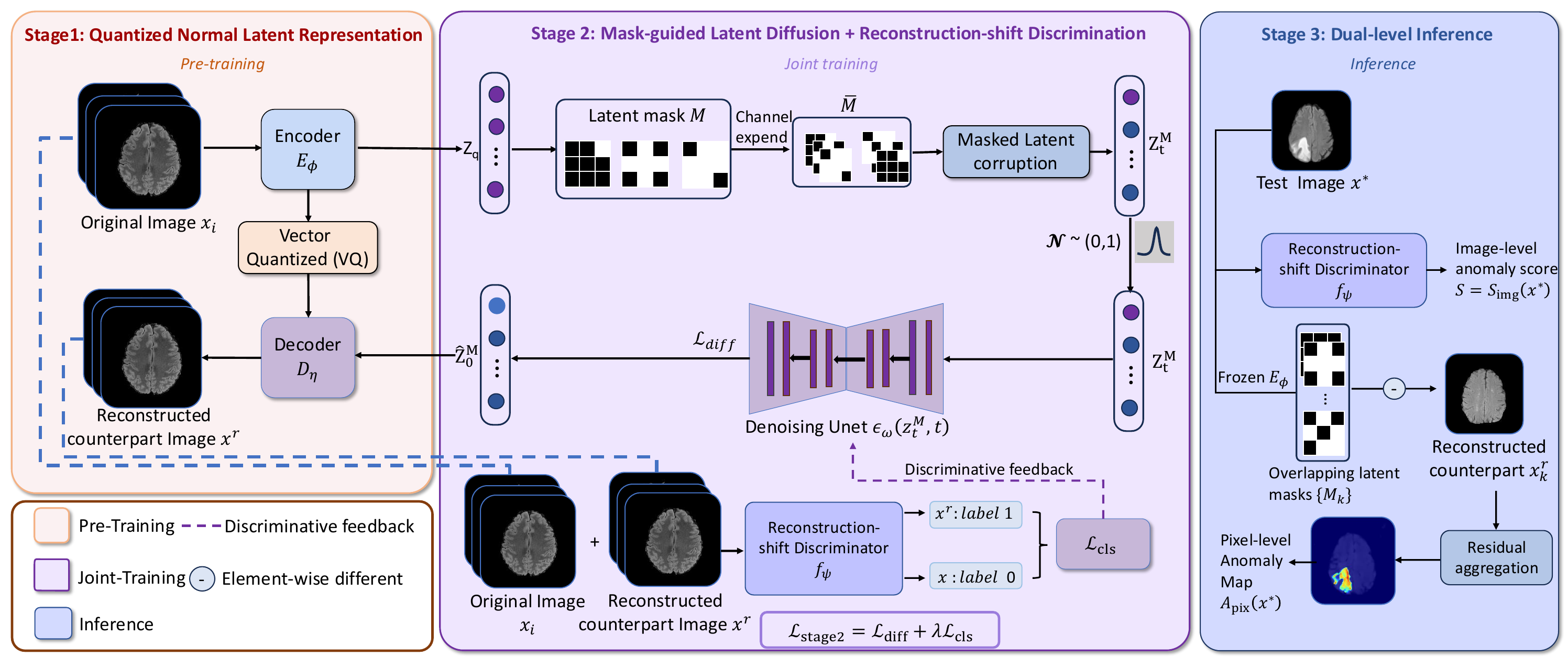}
\caption{Overview of DMD. In Stage I, a vector-quantized encoder--decoder learns a compact normal latent representation. In Stage II, localized masks perturb selected latent positions, which are restored by a diffusion-trained denoiser. The resulting reconstructed counterparts and their original normal images define the reconstruction-shift discrimination task, whose gradients provide discriminative feedback to the denoiser. At inference, the classifier directly produces the image-level anomaly score, while residuals from overlapping masked restorations are aggregated to obtain the pixel-level anomaly map.}
\label{fig1}
\end{figure*}

To answer this question, we propose Discriminative Mask-guided Diffusion (DMD), as illustrated in Figure~\ref{fig1}. DMD is trained in two stages. First, a vector-quantized encoder-decoder learns a compact latent representation from normal images. Second, random spatial masks perturb selected latent regions, and a latent diffusion model reconstructs the perturbed representations. Applying the mask in the quantized latent representation limits the initial perturbation to selected spatial regions rather than altering the full image. Each reconstructed image is paired with its original normal input to define a self-supervised classification task, which we call reconstruction-shift discrimination. This task exposes the classifier to controlled reconstruction changes obtained entirely from normal images. The diffusion and classification objectives are optimized during the same training stage. At inference, DMD provides complementary outputs for image-level detection and pixel-level localization. The classifier assigns an image-level anomaly score to the test image. In parallel, the encoder and latent diffusion model reconstruct the test image according to the patterns learned from normal training data. The absolute difference between the input and its reconstruction forms a pixel-level anomaly map. A single restoration process therefore supports two complementary decisions. The learned classifier supports whole-image detection, while the reconstruction residual preserves spatial information for localization. We summarize the main contributions of this work as follows:
\begin{itemize}[leftmargin=*]
\item We identify reconstruction-shift discrimination, a normal-only self-supervised signal that learns an image-level score from masked diffusion reconstructions and complements residual-based localization.
\item We propose discriminative mask-guided diffusion (DMD), where localized latent masking and latent diffusion let a single restoration process drive both image-level detection and pixel-level localization.
\item We demonstrate that DMD achieves the best overall performance among the compared state-of-the-art baseline methods on five publicly available medical imaging datasets.
\end{itemize}

\section{Related Work}
\label{sec:related_work}

\subsection{Reconstruction-Based Anomaly Detection}
Normal-only medical anomaly detection learns anatomical patterns from normal training images and identifies deviations at test time. Reconstruction-based methods measure input-reconstruction differences using autoencoders, variational autoencoders, adversarial autoencoders, and generative adversarial networks~\cite{atlason2019unsupervised,RN12,makhzani2015adversarial,RN8}. VQ-VAE-based methods~\cite{pinaya2022unsupervised} further organize normal representations through a learned codebook. Residuals preserve spatial information, but their magnitude reflects both pathology and reconstruction quality. Limited-capacity models may produce diffuse errors over normal tissue, whereas expressive models may reproduce pathological structures~\cite{baur2019deep,RN12}. Residuals are therefore useful for localization but less reliable as a single image-level score. Denoising methods recover images from controlled corruption, whose design and available context affect performance~\cite{kascenas2023role}. IterMask2~\cite{liang2024itermask} combines spatial and frequency masking, while IterMask3D~\cite{liang2025itermask3d} extends iterative mask refinement to volumetric brain MRI. These methods motivate localized restoration, but whole-image detection still requires residual aggregation or a dedicated image-level score.

\subsection{Diffusion-Based Anomaly Detection}
Diffusion models recover normal images from noisy inputs and use the reconstructions for anomaly detection. AnoDDPM~\cite{wyatt2022anoddpm} applies partial diffusion with simplex noise. AutoDDPM~\cite{bercea2023generalizing} estimates candidate anomaly regions from diffusion likelihood maps, combines the input with its reconstruction, and resamples the result for harmonized inpainting. Subsequent methods refine restoration through temporal guidance~\cite{bercea2024diffusion}, counterfactual reconstruction~\cite{fontanella2024diffusion}, and patch-based processing~\cite{behrendt2024patched}. cDDPM~\cite{behrendt2025guided} conditions denoising on an input-derived representation to preserve image-specific context. These methods commonly derive spatial anomaly evidence from input-reconstruction differences. Latent diffusion reduces computational cost by operating on a compressed representation~\cite{rombach2022high,iqbal2023unsupervised,pinaya2022fast}. MAD-AD~\cite{beizaee2025mad} is closely related to DMD because it also adds noise to selected latent patches under normal-only training. It jointly identifies corrupted patches and recovers their features, then uses patch predictions during test-time restoration. DMD instead decodes restored latents into full-image counterparts and trains a whole-image reconstruction-shift classifier from original-reconstruction pairs. At test time, the classifier produces the image-level score, while residuals from overlapping masked restorations provide pixel-level localization.

\subsection{Discriminative Anomaly Detection}
Self-supervised discriminative methods construct proxy tasks from transformed normal images. In general anomaly detection, CutPaste~\cite{li2021cutpaste} classifies normal images and cut-and-paste transformations, then applies one-class density estimation for anomaly scoring. DRAEM~\cite{zavrtanik2021draem} trains reconstructive and pixel-level discriminative networks using simulated anomalies and their masks. NSA~\cite{schluter2022natural} creates localized synthetic changes through Poisson image editing. In medical imaging, masked autoencoders have also been combined with classifiers trained on intensity-altered pseudo-abnormal images~\cite{georgescu2023masked}. These methods provide proxy supervision without real abnormal training examples. Discriminative signals can also guide diffusion reconstruction. Classifier-guided DDIM is weakly supervised and uses healthy and diseased image-level labels together with classifier gradients to steer reverse diffusion toward a healthy class~\cite{RN14}. DMD uses only normal images for model optimization and applies no classifier gradients during test-time restoration. It trains the classifier on original normal images and decoded masked-diffusion reconstructions. The resulting whole-image score represents reconstruction shift rather than disease probability, while localization remains residual-based.

\section{Methodology}

\subsection{Problem Formulation}
Let $\mathcal{D}_{N}=\{x_i\}_{i=1}^{N}$ be a training set that contains only normal medical images, where $x_i\in\mathcal{X}$ and $\mathcal{X}=\mathbb{R}^{H\times W\times C}$. Given a test image $x^{*}\in\mathcal{X}$, the goal is to predict an image-level anomaly score $S_{\mathrm{img}}(x^{*})\in[0,1]$ and a pixel-level anomaly map $A_{\mathrm{pix}}(x^{*})\in\mathbb{R}_{+}^{H\times W}$. Higher values indicate stronger deviations from the normal training data.

DMD learns the two outputs from complementary signals. A classifier learns an image-level score by distinguishing original normal images from their diffusion-based reconstructions. The input-reconstruction residual retains spatial information for pixel-level localization. DMD is trained in two stages. The first stage learns a quantized latent representation from normal images. The second stage trains mask-guided latent diffusion together with reconstruction-shift discrimination.

\subsection{Quantized Normal Latent Representation}

DMD performs masking and denoising on a compact spatial latent grid. We first train a vector-quantized autoencoder to map normal images to this grid and reconstruct them from a shared set of code embeddings. The resulting quantized grid provides the operating space for the subsequent masking and diffusion operations. Let $E_{\phi}$ and $D_{\eta}$ denote the encoder and decoder. Given an image $x$, the encoder produces
\begin{equation}
z_e=E_{\phi}(x),
\qquad
z_e\in\mathbb{R}^{h\times w\times d_z}.
\label{eq:encoder}
\end{equation}
Let $\mathcal{C}=\{e_k\}_{k=1}^{K}$ be a learnable codebook, where each $e_k\in\mathbb{R}^{d_z}$. Each spatial latent vector is replaced by its nearest codebook entry:
\begin{equation}
k_{ij}^{*}
=
\underset{k\in\{1,\ldots,K\}}{\arg\min}
\left\|z_e[i,j]-e_k\right\|_2^2,
\qquad
z_q[i,j]=e_{k_{ij}^{*}}.
\label{eq:quantization}
\end{equation}

We use the straight-through estimator to pass the reconstruction gradient to the encoder:
\begin{equation}
z_q^{\mathrm{st}}
=
z_e+\operatorname{sg}\left[z_q-z_e\right],
\qquad
\hat{x}_{\mathrm{vq}}=D_{\eta}\left(z_q^{\mathrm{st}}\right),
\label{eq:straight_through}
\end{equation}
where $\operatorname{sg}[\cdot]$ denotes the stop-gradient operator. The vector-quantized autoencoder is trained with
\begin{equation}
\begin{aligned}
\mathcal{L}_{\mathrm{VQ}}
=
\mathbb{E}_{x\sim\mathcal{D}_{N}}
\big[
&\left\|x-\hat{x}_{\mathrm{vq}}\right\|_2^2
+\left\|\operatorname{sg}[z_e]-z_q\right\|_2^2 \\
&+\beta_{\mathrm{c}}
\left\|z_e-\operatorname{sg}[z_q]\right\|_2^2
\big],
\end{aligned}
\label{eq:vq_loss}
\end{equation}
where $\beta_{\mathrm{c}}$ controls the commitment term. After this stage, $E_{\phi}$, $D_{\eta}$, and $\mathcal{C}$ are fixed to keep the latent representation and decoder stable during Stage II. Diffusion operates on the quantized code embeddings $z_0=z_q$ rather than on the discrete code indices. We next describe how DMD perturbs and restores selected positions on this fixed latent grid.

\subsection{Mask-Guided Latent Diffusion}

To construct localized latent restoration tasks, DMD adds noise only to selected positions and keeps the remaining latent positions unchanged. The retained positions provide clean context for denoising.

For each $z_0\in\mathbb{R}^{h\times w\times d_z}$, we sample a nonempty square mask $M\in\{0,1\}^{h\times w\times 1}$, where one marks a position selected for restoration. Its area ratio is sampled uniformly from $[\rho_{\min},\rho_{\max}]$, and its location is sampled uniformly over the valid latent positions. The mask is copied across the $d_z$ channels and is denoted by $\bar{M}$. The same mask defines the support of both the forward corruption and the diffusion loss.

Let $\{\beta_t\}_{t=1}^{T_d}$ be a fixed noise schedule. We define $\alpha_t=1-\beta_t$ and $\bar{\alpha}_t=\prod_{s=1}^{t}\alpha_s$. For $t$ sampled uniformly from $\{1,\ldots,T_d\}$ and $\epsilon\sim\mathcal{N}(0,I)$, the masked noisy latent is
\begin{equation}
z_t^{M}
=
(1-\bar{M})\odot z_0
+\bar{M}\odot
\left(
\sqrt{\bar{\alpha}_t}z_0
+\sqrt{1-\bar{\alpha}_t}\epsilon
\right).
\label{eq:masked_forward}
\end{equation}
The denoising network $\epsilon_{\omega}$ predicts the injected noise from this partially noised latent. The diffusion objective is evaluated only over the selected positions:
\begin{equation}
\mathcal{L}_{\mathrm{diff}}
=
\mathbb{E}_{x,M,t,\epsilon}
\left[
\frac{
\left\|
\bar{M}\odot
\left(
\epsilon-\epsilon_{\omega}(z_t^{M},t)
\right)
\right\|_2^2
}{d_z\left\|M\right\|_1}
\right].
\label{eq:masked_diffusion_loss}
\end{equation}

A single denoiser evaluation provides an estimate of the clean latent at timestep $t$. The unmasked positions are copied directly from $z_0$:
\begin{equation}
\hat{z}_0^{M}
=
(1-\bar{M})\odot z_0
+\bar{M}\odot
\frac{
z_t^{M}
-\sqrt{1-\bar{\alpha}_t}\epsilon_{\omega}(z_t^{M},t)
}{\sqrt{\bar{\alpha}_t}}.
\label{eq:clean_latent_estimate}
\end{equation}
The fixed decoder maps this estimate back to the image space:
\begin{equation}
x^{r}=D_{\eta}\left(\hat{z}_0^{M}\right),
\label{eq:reconstructed_counterpart}
\end{equation}
where $x^{r}$ is the reconstructed counterpart of $x$. The mask confines the initial corruption to selected latent positions, while the fixed decoder maps the restored latent back to the full image. This reconstructed counterpart also provides the positive example for the image-level task described next. More latent mask rule are provided in Appendix~\ref{app:latent_mask}.

\subsection{Reconstruction-Shift Discrimination}

The restoration process above produces two paired views of each normal image, namely the original image $x$ and its reconstructed counterpart $x^r$. Their residual preserves spatial information, but reducing it to one scalar removes that spatial structure and can mix errors from different regions. DMD therefore uses the same pair to define a normal-only classification task for image-level detection.

Let $f_{\psi}(u)\in[0,1]$ denote the probability assigned to the reconstruction-shift class. Original normal images are assigned label zero, and their reconstructed counterparts are assigned label one. For a batch of $B$ normal images, the classification loss is
\begin{equation}
\mathcal{L}_{\mathrm{cls}}
=
-\frac{1}{2B}
\sum_{i=1}^{B}
\left[
\log\left(1-f_{\psi}(x_i)\right)
+\log f_{\psi}\left(x_i^{r}\right)
\right].
\label{eq:classification_loss}
\end{equation}
The positive label denotes a reconstruction-induced shift rather than a disease category. Both labels are therefore available from normal training data. The classifier learns to distinguish original normal images from outputs of the masked restoration process.

\subsection{Two-Stage Optimization}

We separate latent representation learning from reconstruction-shift learning so that the second stage operates on a stable representation and decoder. In the first stage, $E_{\phi}$, $D_{\eta}$, and $\mathcal{C}$ are optimized using $\mathcal{L}_{\mathrm{VQ}}$. These components are fixed in the second stage, while the denoising network and classifier are optimized with
\begin{equation}
\mathcal{L}_{\mathrm{stage2}}
=
\mathcal{L}_{\mathrm{diff}}
+\lambda\mathcal{L}_{\mathrm{cls}},
\label{eq:stage2_loss}
\end{equation}
where $\lambda$ balances masked diffusion denoising and reconstruction-shift discrimination. The diffusion loss trains the denoiser to recover the selected latent positions. The reconstructed counterpart in Eq.~\eqref{eq:reconstructed_counterpart} remains differentiable, so $\mathcal{L}_{\mathrm{cls}}$ updates $f_{\psi}$ and also passes through the fixed decoder to $\epsilon_{\omega}$. Since $x^r$ belongs to class one, this gradient favors reconstruction shifts that remain distinguishable from their original inputs. The diffusion objective constrains this update through masked noise prediction on normal latent representations. After training, the classifier and restoration branches provide the two anomaly outputs described next. More two-stage training are provided in Appendix~\ref{app:two_stage_training}.

\subsection{Dual-Level Anomaly Inference}

Given a test image $x^{*}$, the classifier branch does not require reconstruction and directly provides
\begin{equation}
S_{\mathrm{img}}(x^{*})=f_{\psi}(x^{*}).
\label{eq:image_score}
\end{equation}
We use this reconstruction-shift probability as the image-level anomaly score. A larger value indicates greater similarity to the shift class learned from normal data.

The classifier score has no spatial output, so we use the restoration branch for pixel-level localization. We encode and quantize $x^{*}$ to obtain $z_0^{*}$. Overlapping square masks $\{M_k\}_{k=1}^{K_M}$ with side length $\ell_M$ are placed on a regular grid with stride $s_M$ to cover every latent position. For each mask, we draw one Gaussian noise sample at a fixed restoration step $t^{*}$ and apply Eq.~\eqref{eq:masked_forward} to obtain $z_{t^{*}}^{M_k}$. A single denoiser evaluation then produces $\hat{z}_0^{M_k}$ through Eq.~\eqref{eq:clean_latent_estimate}, which is decoded as
\begin{equation}
x_k^{r}=D_{\eta}\left(\hat{z}_0^{M_k}\right).
\label{eq:test_reconstruction}
\end{equation}
Training and inference use the same masked corruption equation and clean-latent estimator. Training samples masks and timesteps randomly, while inference uses grid masks and a fixed $t^{*}$ for full spatial coverage.

Each mask produces one reconstructed counterpart and its associated residual map. Let $M_k^{\uparrow}\in\{0,1\}^{H\times W}$ denote the image-space mask obtained by nearest-neighbor upsampling. The channel-averaged residual for the $k$th reconstruction is
\begin{equation}
R_k(i,j)
=
\frac{1}{C}
\sum_{c=1}^{C}
\left|
x_{ijc}^{*}-x_{k,ijc}^{r}
\right|.
\label{eq:residual_map}
\end{equation}
Because the grid masks overlap, a pixel can be included in more than one restoration task. The final pixel-level anomaly map averages the residuals over all masks that cover each pixel:
\begin{equation}
A_{\mathrm{pix}}(x^{*})_{ij}
=
\frac{
\sum_{k=1}^{K_M}M_k^{\uparrow}(i,j)R_k(i,j)
}{
\sum_{k=1}^{K_M}M_k^{\uparrow}(i,j)+\delta
},
\label{eq:anomaly_map}
\end{equation}
where $\delta$ is a small constant for numerical stability. The
training mask range $[\rho_{\min},\rho_{\max}]$ and the inference
parameters $\ell_M$, $s_M$, and $t^{*}$ are selected using only normal training data and are fixed before test evaluation. Together, the classifier and restoration branches provide complementary predictions for image-level detection and pixel-level localization. Additional implementation details of dual-level inference are provided in Appendix~\ref{app:dual_level_inference}.

\begin{table*}[t]
\centering
\setlength{\tabcolsep}{2mm}
\resizebox{\textwidth}{!}{%
\begin{tabular}{@{}lccccccccc@{}}
\toprule
\multirow{2}{*}{\textbf{Method}}
& \multicolumn{3}{c}{\textbf{BraTS2021}}
& \multicolumn{3}{c}{\textbf{BUSI}}
& \multicolumn{3}{c}{\textbf{VinDr-CXR}} \\
\cmidrule(lr){2-4}
\cmidrule(lr){5-7}
\cmidrule(lr){8-10}
& AUC$\uparrow$ & AP$\uparrow$ & F1$\uparrow$
& AUC$\uparrow$ & AP$\uparrow$ & F1$\uparrow$
& AUC$\uparrow$ & AP$\uparrow$ & F1$\uparrow$ \\
\midrule

AE
& \ms{82.6}{0.37} & \ms{92.0}{0.56} & \ms{89.4}{0.25}
& \ms{86.1}{0.42} & \ms{84.6}{0.41} & \ms{81.3}{0.36}
& \ms{56.4}{0.40} & \ms{60.2}{0.45} & \ms{57.3}{0.36} \\

VAE
& \ms{80.6}{0.29} & \ms{90.9}{0.37} & \ms{88.6}{0.26}
& \ms{84.2}{0.33} & \ms{83.3}{0.45} & \ms{80.1}{0.32}
& \ms{56.5}{0.35} & \ms{60.1}{0.42} & \ms{57.5}{0.29} \\

f-AnoGAN
& \ms{85.1}{0.28} & \ms{93.5}{0.45} & \ms{91.0}{0.14}
& \ms{86.7}{0.35} & \ms{85.2}{0.42} & \ms{82.3}{0.25}
& \ms{74.7}{0.33} & \ms{72.8}{0.41} & \ms{73.0}{0.27} \\

DAE
& \ms{85.1}{0.23}
& \ms{93.5}{0.05}
& \underline{\ms{92.8}{0.41}}
& \ms{87.2}{0.65}
& \ms{86.5}{0.69}
& \ms{85.5}{0.12}
& \ms{68.7}{0.42}
& \ms{71.8}{0.24}
& \ms{75.9}{0.13} \\

\midrule

AnoDDPM
& \ms{82.7}{0.16} & \ms{91.7}{0.26} & \ms{89.9}{0.05}
& \ms{86.8}{0.21} & \ms{84.0}{0.32} & \ms{85.2}{0.18}
& \ms{41.2}{0.25} & \ms{42.5}{0.30} & \ms{44.8}{0.26} \\

AutoDDPM
& \ms{85.6}{0.47} & \ms{93.1}{0.25} & \ms{91.1}{0.44}
& \ms{89.2}{0.38} & \ms{85.0}{0.27} & \ms{86.7}{0.30}
& \ms{69.4}{0.29} & \ms{68.6}{0.34} & \ms{68.4}{0.24} \\

THOR$_{\text{Gaussian}}$
& \ms{85.9}{0.16} & \ms{93.4}{0.45} & \ms{91.5}{0.14}
& \ms{89.6}{0.23} & \ms{85.4}{0.32} & \ms{88.1}{0.25}
& \ms{74.4}{0.31} & \ms{75.4}{0.33} & \ms{74.7}{0.26} \\

THOR$_{\text{Simplex}}$
& \underline{\ms{86.1}{0.26}}
& \underline{\ms{93.7}{0.05}}
& \ms{91.8}{0.04}
& \underline{\ms{89.9}{0.31}}
& \ms{85.7}{0.25}
& \ms{89.3}{0.20}
& \ms{75.3}{0.27}
& \ms{76.5}{0.30}
& \ms{75.8}{0.23} \\

Dif-fuse
& \ms{83.0}{0.37} & \ms{92.6}{0.16} & \ms{91.7}{0.24}
& \ms{87.6}{0.28} & \ms{84.5}{0.22} & \ms{90.0}{0.17}
& \ms{79.8}{0.30} & \ms{78.7}{0.32} & \ms{78.5}{0.24} \\

cDDPM
& \ms{85.4}{0.26} & \ms{93.0}{0.25} & \ms{91.0}{0.14}
& \ms{89.1}{0.34} & \ms{88.9}{0.29} & \ms{91.5}{0.19}
& \ms{79.3}{0.28} & \ms{80.4}{0.30} & \ms{80.0}{0.24} \\

MAD-AD
& \ms{85.7}{0.06}
& \ms{93.2}{0.15}
& \ms{91.3}{0.24}
& \ms{89.4}{0.19}
& \underline{\ms{91.1}{0.20}}
& \underline{\ms{91.8}{0.22}}
& \underline{\ms{81.3}{0.22}}
& \underline{\ms{83.2}{0.24}}
& \underline{\ms{82.6}{0.18}} \\

\midrule

\rowcolor{blue!10}
\textbf{DMD}
& \textbf{\ms{87.2}{0.25}}
& \textbf{\ms{94.3}{0.14}}
& \textbf{\ms{94.4}{0.13}}
& \textbf{\ms{93.8}{0.27}}
& \textbf{\ms{96.0}{0.18}}
& \textbf{\ms{93.9}{0.20}}
& \textbf{\ms{84.3}{0.18}}
& \textbf{\ms{84.2}{0.21}}
& \textbf{\ms{83.9}{0.15}} \\

\bottomrule
\end{tabular}
}
\caption{Image-level anomaly detection performance on BraTS2021, BUSI, and VinDr-CXR averaged over five runs with mean±std. The best and second-best results are marked in \textbf{bold} and \underline{underline}, respectively.}
\label{tab:image-level}
\end{table*}

\section{Experiment}
\subsection{Experimental Setup}
\paragraph{Datasets}
We evaluate DMD on five publicly available medical image datasets covering brain MRI, breast ultrasound, and chest radiography: BraTS2021, IXI/ATLAS 2.0, WMH, BUSI, and VinDr-CXR. For BraTS2021 and VinDr-CXR, we follow the preprocessing and train/test partition protocol established by MedIAnomaly~\cite{cai2025medianomaly} to ensure comparability with prior medical anomaly detection benchmarks. Detailed dataset descriptions are provided in Appendix~\ref{app:datasets}.

\paragraph{Baseline Methods}
We compare our method with classical reconstruction-based methods, including: (1) four classical methods: AE~\cite{atlason2019unsupervised}, VAE~\cite{RN12}, f-AnoGAN~\cite{RN8} and DAE~\cite{kascenas2023role}, and (2) seven diffusion-based methods: DDPM \cite{ho2020denoising}, AnoDDPM \cite{wyatt2022anoddpm}, AutoDDPM \cite{bercea2023generalizing}, THOR \cite{bercea2024diffusion}, Dif-fuse \cite{fontanella2024diffusion}, cDDPM \cite{behrendt2025guided}, and MAD-AD \cite{beizaee2025mad}.

\paragraph{Evaluation Metrics}
We evaluate DMD at both image and pixel levels. At the image level, we report AUC, AP, and F1. At the pixel level, we report AP$_{\mathrm{pix}}$ and Dice. AUC, AP, and AP$_{\mathrm{pix}}$ are threshold-free metrics, whereas the thresholds for F1 and Dice are calibrated exclusively using normal training data without access to test labels. For brain imaging datasets, we additionally report RQI, AHI, and CACI following Bercea et al.~\cite{bercea2025evaluating}. Detailed metric definitions are provided in Appendix~\ref{app:implementation_details}.

\paragraph{Implementation Details.}
All images are center-cropped, resized to $128{\times}128$, and normalized to $[-1,1]$. Training uses only normal images, while testing includes both normal and abnormal samples. Subject-level splits are used whenever subject identifiers are available. DMD is optimized in two stages. We pretrain the VQ-VAE for 250 epochs, then fix its encoder, decoder, and codebook. The latent denoising network and reconstruction-shift classifier are jointly trained for 300 epochs in the second stage. Unless otherwise specified, we use latent dimension $d_z=64$, codebook size $K=256$, diffusion horizon $T_d=1000$, and classification weight $\lambda=0.1$. We use AdamW with a learning rate of $2\times10^{-4}$ and a batch size of 22. All experiments are conducted on a single NVIDIA A100 GPU with 80 GB of memory. Additional network and inference details are provided in Appendix~\ref{app:implementation_details}.

\subsection{Comparison with State-of-the-Art Methods}
\paragraph{Image-level Medical Anomaly Detection}
Table~\ref{tab:image-level} reports the image-level anomaly detection results on BraTS2021, BUSI, and VinDr-CXR. DMD consistently ranks first across all datasets and metrics, indicating robust image-level anomaly discrimination across MRI, ultrasound, and chest X-ray modalities. For the primary threshold-free metric AUC, DMD achieves 87.2\%, 93.8\%, and 84.3\% on the three datasets, improving over the strongest competing method by 1.1\%, 3.9\%, and 3.0\%, respectively. DMD also obtains the highest F1 scores, with gains of 3.1\%, 2.1\%, and 1.3\% over MAD-AD. These improvements show that DMD produces more discriminative image-level anomaly scores while maintaining consistent performance across different medical imaging domains.

\begin{table*}[ht!]
\centering
\setlength{\tabcolsep}{1.6mm}
\resizebox{\textwidth}{!}{%
\begin{tabular}{@{}lcccccccc@{}}
\toprule
\multirow{2}{*}{\textbf{Method}} &
\multicolumn{2}{c}{\textbf{BraTS2021}} &
\multicolumn{2}{c}{\textbf{ATLAS 2.0}} &
\multicolumn{2}{c}{\textbf{WMH}} &
\multicolumn{2}{c}{\textbf{BUSI}}\\
\cmidrule(lr){2-3}
\cmidrule(lr){4-5}
\cmidrule(lr){6-7}
\cmidrule(lr){8-9}
& AP$_{\text{pix}}\uparrow$ & Dice$\uparrow$
& AP$_{\text{pix}}\uparrow$ & Dice$\uparrow$
& AP$_{\text{pix}}\uparrow$ & Dice$\uparrow$
& AP$_{\text{pix}}\uparrow$ & Dice$\uparrow$\\
\midrule

AE \cite{atlason2019unsupervised}
& \ms{33.2}{1.62} & \ms{39.2}{1.24}
& \ms{7.5}{0.48} & \ms{13.5}{1.41}
& \ms{12.3}{1.32} & \ms{13.7}{1.22}
& \ms{50.4}{1.63} & \ms{53.8}{1.45}\\

VAE \cite{RN12}
& \ms{44.0}{1.71} & \ms{47.1}{1.54}
& \ms{10.2}{0.39} & \ms{15.6}{1.14}
& \ms{13.1}{1.42} & \ms{14.6}{1.22}
& \ms{58.9}{1.51} & \ms{62.1}{1.35}\\

f-AnoGAN \cite{RN8}
& \ms{23.7}{1.25} & \ms{28.5}{1.14}
& \ms{11.3}{0.35} & \ms{16.9}{1.04}
& \ms{15.5}{1.21} & \ms{17.2}{1.41}
& \ms{55.7}{1.44} & \ms{60.2}{1.50}\\

DAE \cite{kascenas2023role}
& \underline{\ms{74.9}{0.24}}
& \underline{\ms{70.2}{0.52}}
& \ms{13.4}{0.04}
& \ms{17.8}{0.12}
& \ms{38.2}{0.34}
& \ms{45.1}{0.18}
& \ms{62.3}{0.34}
& \ms{64.2}{0.21}\\

\midrule

DDPM \cite{ho2020denoising}
& \ms{52.8}{0.48} & \ms{51.2}{1.16}
& \ms{12.1}{0.49} & \ms{17.6}{1.21}
& \ms{14.8}{1.40} & \ms{17.5}{1.21}
& \ms{63.2}{1.73} & \ms{66.9}{1.54}\\

AnoDDPM \cite{wyatt2022anoddpm}
& \ms{53.4}{1.27} & \ms{51.5}{1.15}
& \ms{12.6}{0.49} & \ms{18.1}{1.11}
& \ms{13.6}{1.10} & \ms{15.1}{1.50}
& \ms{68.1}{1.85} & \ms{69.3}{1.64}\\

AutoDDPM \cite{bercea2023generalizing}
& \ms{55.1}{0.16} & \ms{65.0}{1.14}
& \ms{14.8}{1.30} & \ms{19.9}{1.42}
& \ms{45.3}{1.71} & \ms{50.3}{1.46}
& \ms{69.0}{1.64} & \ms{68.2}{1.52}\\

THOR$_{\text{Gaussian}}$ \cite{bercea2024diffusion}
& \ms{58.7}{1.15} & \ms{69.7}{1.33}
& \ms{15.3}{1.10} & \ms{20.4}{0.21}
& \ms{46.4}{1.36} & \ms{51.6}{0.15}
& \ms{69.4}{1.54} & \ms{70.8}{1.43}\\

THOR$_{\text{Simplex}}$ \cite{bercea2024diffusion}
& \ms{62.4}{1.05} & \ms{64.6}{1.04}
& \ms{24.1}{1.23} & \ms{29.3}{1.13}
& \ms{47.9}{0.46} & \ms{53.2}{1.45}
& \ms{70.2}{1.40} & \ms{65.1}{1.33}\\

Dif-fuse \cite{fontanella2024diffusion}
& \ms{65.8}{1.14} & \ms{67.7}{1.13}
& \ms{25.2}{1.42} & \ms{30.4}{1.30}
& \ms{51.2}{1.15} & \ms{56.9}{1.04}
& \underline{\ms{70.6}{1.33}}
& \underline{\ms{71.8}{1.36}}\\

cDDPM \cite{behrendt2025guided}
& \ms{52.7}{1.18} & \ms{56.3}{1.54}
& \ms{19.0}{1.11} & \ms{24.2}{1.43}
& \ms{52.1}{1.45}
& \underline{\ms{57.6}{1.14}}
& \ms{66.8}{1.67} & \ms{68.5}{1.45}\\

MAD-AD \cite{beizaee2025mad}
& \ms{63.9}{1.15} & \ms{69.6}{1.43}
& \underline{\ms{46.4}{1.14}}
& \underline{\ms{51.6}{1.15}}
& \underline{\ms{53.2}{1.04}}
& \ms{55.7}{1.33}
& \ms{69.8}{1.58} & \ms{71.0}{1.30}\\

\midrule

\rowcolor{blue!10}
\textbf{DMD}
& \textbf{\ms{78.9}{1.13}}
& \textbf{\ms{76.3}{1.42}}
& \textbf{\ms{49.5}{1.31}}
& \textbf{\ms{54.7}{1.32}}
& \textbf{\ms{56.4}{1.14}}
& \textbf{\ms{60.9}{1.33}}
& \textbf{\ms{71.3}{1.48}}
& \textbf{\ms{72.4}{1.30}}\\

\bottomrule
\end{tabular}%
}
\caption{Pixel-level anomaly detection performance on BraTS2021, ATLAS 2.0, WMH, and BUSI averaged over five runs with mean$\pm$std. The best and second-best results are marked in \textbf{bold} and \underline{underline}, respectively.}
\label{tab:pixel-level}
\end{table*}

\paragraph{Pixel-level Medical Anomaly Detection}
Table~\ref{tab:pixel-level} presents the pixel-level anomaly localization results on BraTS2021, ATLAS 2.0, WMH, and BUSI. DMD achieves the highest AP$_\text{pix}$ and Dice scores across all four datasets, demonstrating consistent localization performance on both brain MRI and breast ultrasound images. Compared with classical reconstruction-based methods, DMD shows a advantage, especially on the brain MRI benchmarks where conventional autoencoding or GAN-based models produce substantially weaker lesion localization. More importantly, DMD also outperforms recent diffusion-based baselines. Relative to the strongest competitor MAD-AD, DMD improves AP$_\text{pix}$ by 15.0\%, 3.1\%, and 3.2\% on BraTS2021, ATLAS 2.0, and WMH, respectively, while also achieving higher Dice scores on all three datasets. On BUSI, DMD obtains 71.3\% AP$_\text{pix}$ and 72.4\% Dice, surpassing both classical and diffusion-based baselines. These results indicate that the proposed framework provides more accurate anomaly localization across different anatomical regions and imaging modalities.

The WMH evaluation further tests cross-dataset robustness, where the model trained on BraTS2021 is applied to WMH without retraining. Although both datasets use FLAIR MRI, they differ in lesion type, morphology, and acquisition characteristics. The performance gain on WMH therefore indicates that DMD is not limited to tumor-specific reconstruction, but can generalize to pathology-level distribution shifts. Additional evaluation of normative reconstruction quality using RQI, AHI, and CACI is provided in Appendix~\ref{app:normative_quality}.

\begin{figure}[t]
    \centering
    \includegraphics[width=\linewidth]{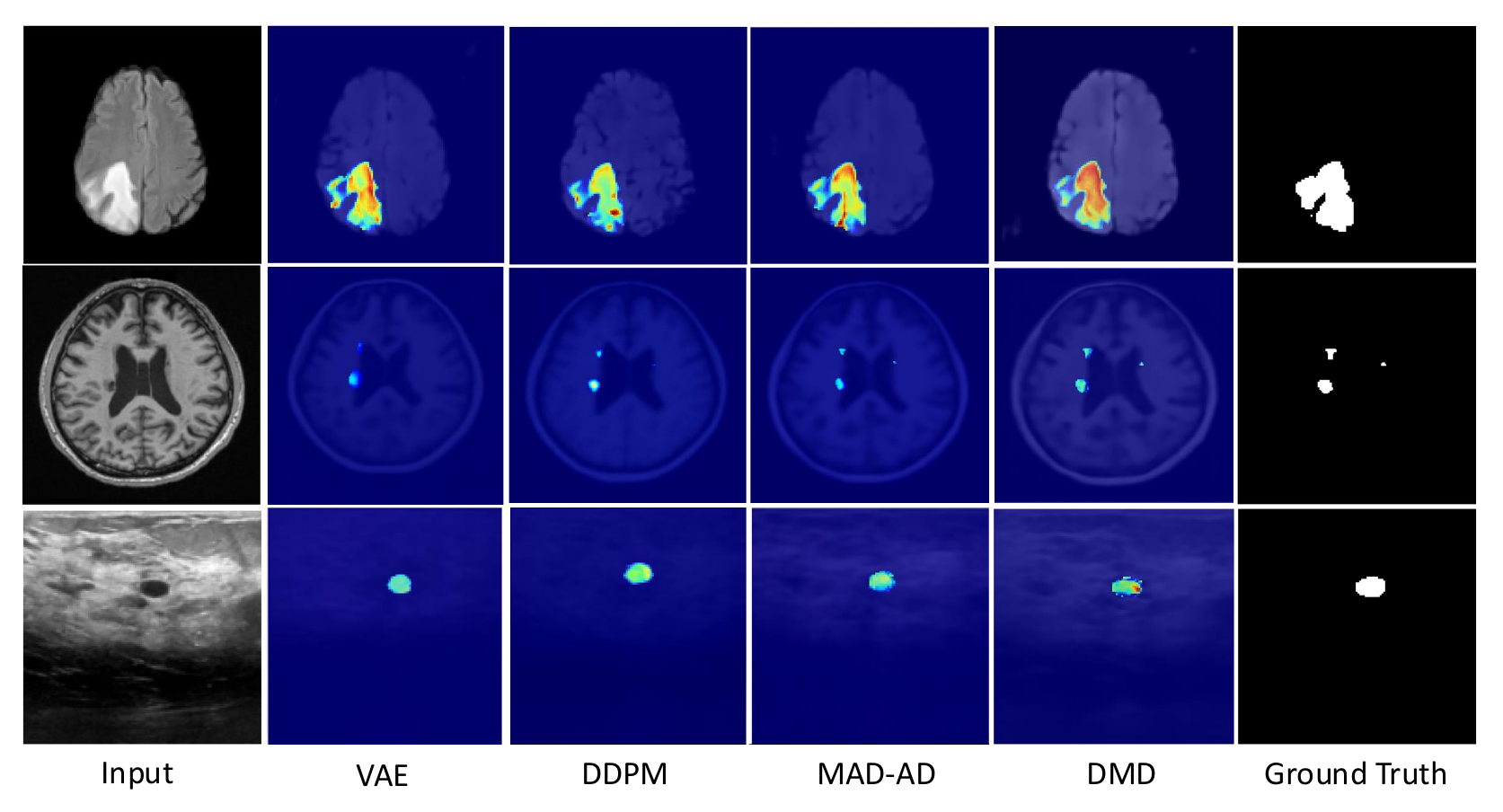}
    \caption{Qualitative comparison of anomaly localization results on BraTS2021, ATLAS 2.0, and BUSI. From left to right, each row shows the input image, anomaly maps generated by VAE, DDPM, MAD-AD, and DMD, and the ground-truth annotation.}
    \label{fig:visual comparison}
\end{figure}

\paragraph{Qualitative Analysis}
To qualitatively assess the localization ability of DMD, we compare its anomaly maps with those of VAE, DDPM, and MAD-AD on representative cases from BraTS2021, ATLAS 2.0, and BUSI, as shown in Figure~\ref{fig:visual comparison}. Each row shows the input, predicted anomaly maps, and ground-truth annotation. On BraTS2021, DMD produces a spatially coherent response that better matches the tumor boundary, whereas the baselines yield less complete or concentrated activations. For the small, low-contrast lesion in ATLAS 2.0, VAE and DDPM produce weak, scattered responses, while DMD localizes the abnormal regions more clearly. DMD also generates a compact response aligned with the BUSI lesion. Overall, these examples suggest that DMD provides sharper and more localized anomaly maps, particularly for subtle abnormalities. Additional visualizations are provided in Appendix~\ref{app:additional_visualizations}.

\begin{figure}[t]
    \centering
    \includegraphics[width=\linewidth]{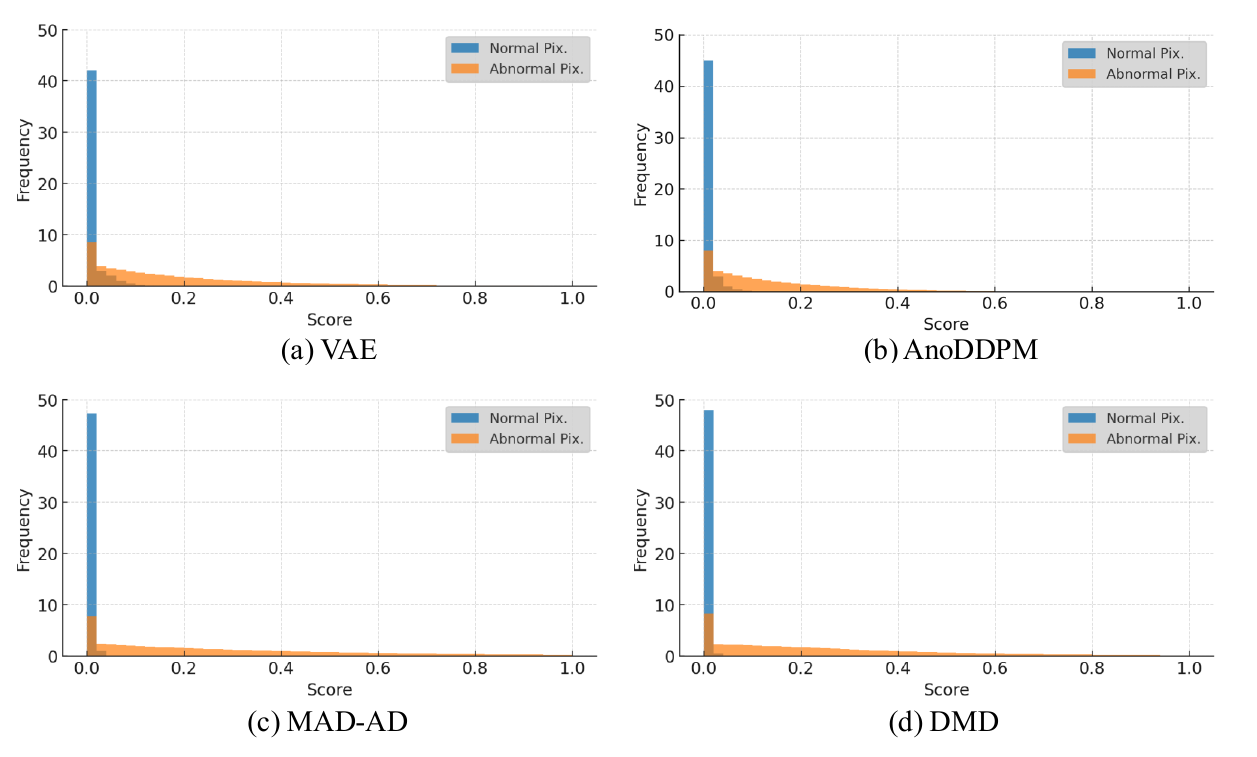}
    \caption{Pixel-level anomaly score distributions on BraTS2021. Blue bars denote normal pixels and orange bars denote abnormal pixels.}
    \label{fig:imgspace-hist}
\end{figure}

\paragraph{Pixel-score Distribution Analysis}
Figure~\ref{fig:imgspace-hist} visualizes the image-space anomaly-score distributions of normal and abnormal pixels on BraTS2021. Across all methods, normal pixels concentrate near zero, as normal anatomy dominates each image and produces small reconstruction residuals. However, VAE and AnoDDPM exhibit overlap between the two distributions, indicating limited discrimination between lesions and normal anatomical variations. MAD-AD reduces this overlap, although many abnormal pixels remain in the low-score region. In contrast, DMD produces a compact normal-pixel distribution near zero and assigns higher scores to abnormal pixels, yielding clearer separation between healthy and pathological regions. Additional parameter-sensitivity analyses are provided in Appendix~\ref{app:parameter_sensitivity}.

\begin{figure}[t]
    \centering
    \begin{subfigure}[t]{0.45\linewidth}
        \centering
        \includegraphics[width=\linewidth, height=0.15\textheight]{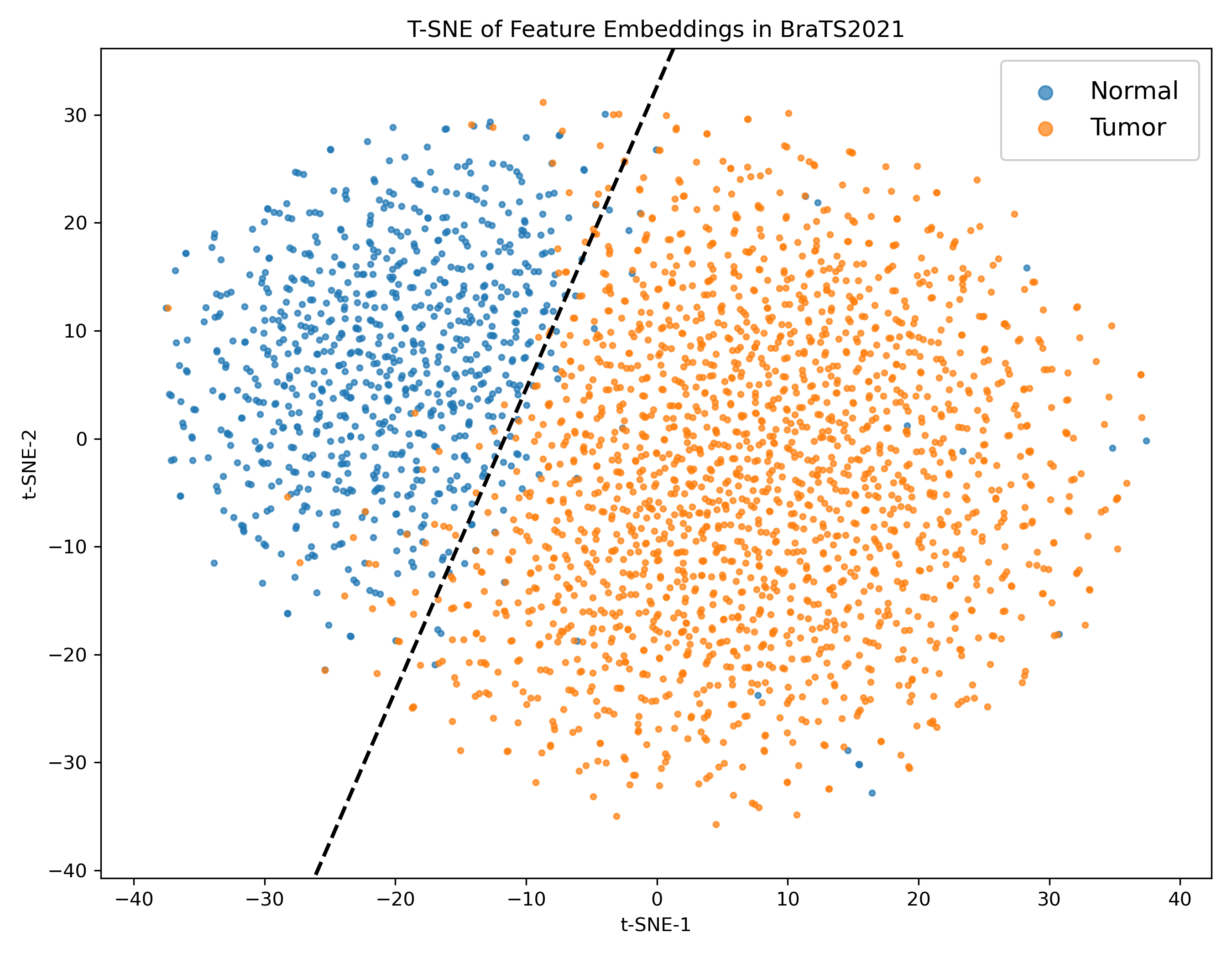}
        \caption{BraTS2021}
        \label{fig:tsne1}
    \end{subfigure}
 \hfill
    \begin{subfigure}[t]{0.45\linewidth}
        \centering
        \includegraphics[width=\linewidth, height=0.15\textheight]{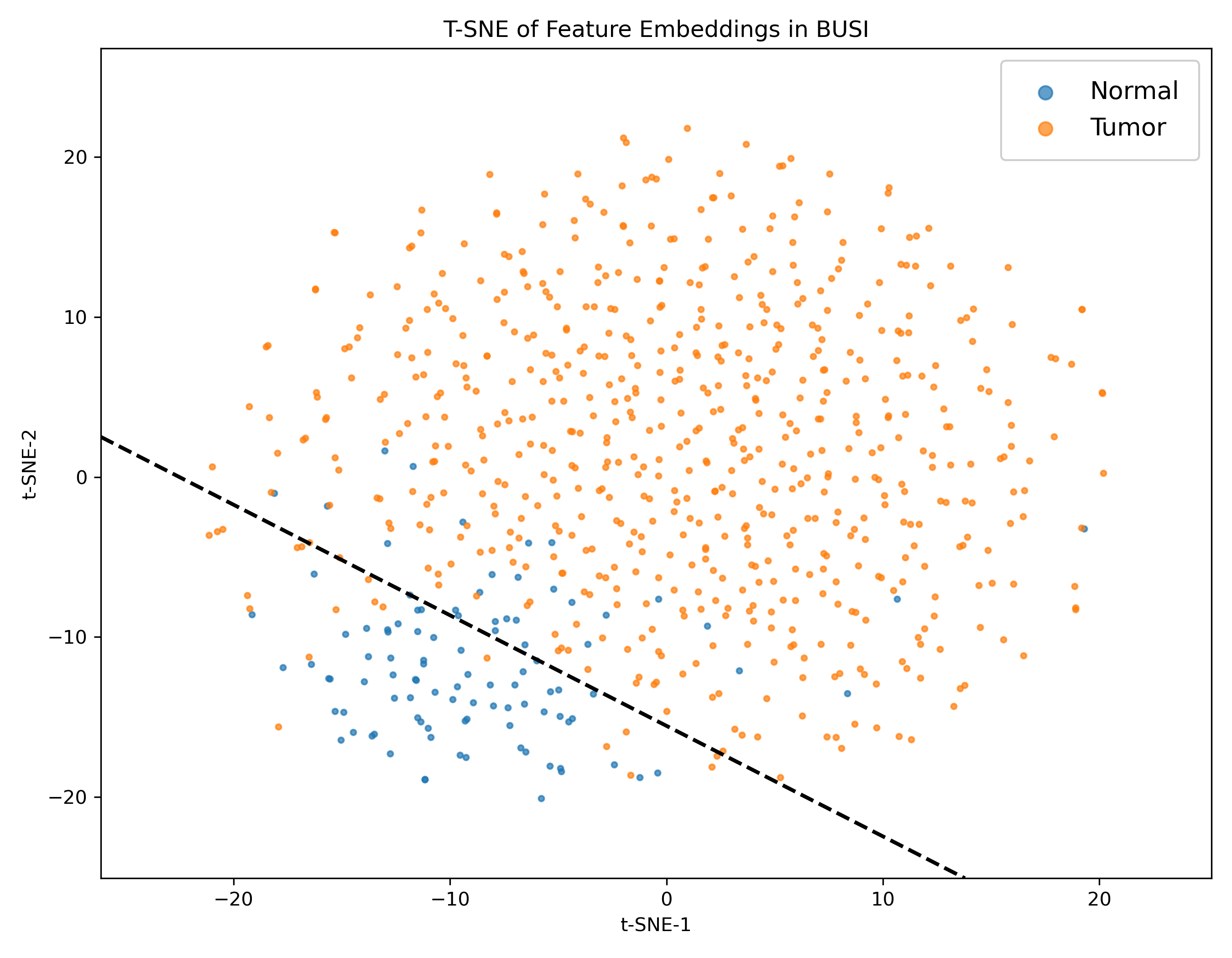}
        \caption{BUSI}
        \label{fig:tsne2}
    \end{subfigure}
    \caption{T-SNE visualization of feature embeddings extracted from the BraTS2021 and BUSI.}
    \label{fig:tsne_vis}
\end{figure}

\paragraph{Visualization of Discriminative Feature Embeddings}
\label{app:tsne_embedding}
We examine whether representations learned through reconstruction-shift discrimination encode pathological deviations. Specifically, we extract image-level features from the trained discriminator $f_{\psi}$ for BraTS2021 and BUSI test images and project them into two dimensions using t-SNE (Figure~\ref{fig:tsne_vis}). Normal and abnormal samples exhibit visible separation, although some overlap remains. The dashed line represents a post-hoc linear SVM fitted in the t-SNE space solely as a visual guide. Together with the quantitative results, these projections suggest that reconstruction-shift discrimination learns anomaly-sensitive representations.

\subsection{Ablation Study}
In this section, we conduct an ablation study to assess the contribution of each component within our proposed framework. Specifically, we construct five variants:
\begin{enumerate}
\item \textbf{w/o Classifier}: We remove the reconstruction-shift discriminator.
\item \textbf{w/o Discriminative Feedback}: We retain and train the reconstruction-shift discriminator, but detach the reconstructed images when computing $\mathcal{L}_{\mathrm{cls}}$.
\item \textbf{w/o Pre-train}: In this variant, the model is directly trained without a pre-training phase.
\item \textbf{w/o Masking Strategy}: In this variant, we remove the mask-based guidance during training.
\item \textbf{w/o Latent Diffusion}: In this variant, we only utilize the autoencoder to generate pseudo-healthy images. Particularly, we preserve the masking strategy.
\end{enumerate}

Table~\ref{tab:ablation_brats} summarizes the results. Replacing the classifier with an aggregated reconstruction-residual score leads to lower image-level performance, showing that the learned discriminator provides a more informative anomaly score than residual magnitude alone. Retaining the classifier while blocking the gradients of $\mathcal{L}_{\mathrm{cls}}$ to the denoising network improves image-level detection but remains inferior to the full model. This comparison isolates the benefit of discriminative feedback, which improves not only the decision boundary but also the reconstructed counterparts used for pixel-level localization. Removing latent diffusion causes the largest degradation, confirming its importance for modeling anatomical variations and producing plausible reconstructions. Removing latent masking particularly reduces localization performance, indicating that localized corruption encourages spatially targeted restoration. Finally, training without VQ-VAE pretraining moderately degrades performance, suggesting that the pretrained quantized representation provides a stable normal latent space for subsequent joint optimization.

\begin{table}[!t]
\centering
\footnotesize
\setlength{\tabcolsep}{1.7mm}
\begin{tabular}{l|cc|cc}
\toprule
\multirow{2}{*}{\textbf{Variant}}
& \multicolumn{2}{c|}{\textbf{Image-level}}
& \multicolumn{2}{c}{\textbf{Pixel-level}} \\
\cmidrule{2-5}
& \textbf{AP} & \textbf{F1}
& \textbf{AP$_{\mathrm{pix}}$} & \textbf{Dice} \\
\midrule
w/o classifier
    & 89.7 & 90.5 & 72.8 & 71.6 \\
w/o Discriminative Feedback
    & 92.1 & 92.7 & 72.9 & 71.7 \\
w/o Pre-training
    & 88.5 & 89.3 & 56.8 & 62.5 \\
w/o Masking Strategy
    & 90.6 & 91.5 & 55.2 & 65.8 \\
w/o Latent Diffusion
    & 87.4 & 88.6 & 44.8 & 47.9 \\
\midrule
DMD
    & \textbf{94.3} & \textbf{94.4}
    & \textbf{78.9} & \textbf{76.3} \\
\bottomrule
\end{tabular}
\caption{Component ablation on BraTS2021.}
\label{tab:ablation_brats}
\end{table}

\section{Conclusion}
In this paper, we propose the discriminative mask-guided diffusion (DMD) framework for image-level anomaly detection and pixel-level localization. DMD addresses the ambiguity of using a single reconstruction residual for both tasks. Specifically, DMD applies localized perturbations to a quantized latent representation and restores the selected regions with a diffusion-trained denoiser. The resulting reconstruction shifts provide proxy supervision for image-level discrimination, while residuals from the same restoration process support pixel-level localization. Across five publicly available medical imaging benchmarks, DMD achieves the best overall performance among the compared baseline methods. Overall, DMD complements residual-based localization with a learned image-level signal derived entirely from normal training data. A current limitation is that the image-level signal is learned from reconstruction shifts produced by a fixed masking policy. How well this proxy transfers across unseen pathologies and acquisition conditions remains to be studied more broadly. Future work will explore adaptive shift construction, multi-center evaluation, and extension to volumetric imaging.

\bibliography{dmd}

\clearpage
\appendix
\setcounter{equation}{0}
\renewcommand{\theequation}{A\arabic{equation}}
\section{Detailed Algorithms of DMD}
\label{app:dmd_algorithms}

The main paper defines the model and objectives. This section provides the exact mask rule and execution order used for training and inference without repeating the main equations.

\subsection{Latent Mask Construction}
\label{app:latent_mask}

For each training sample, we draw an area ratio $\rho\sim\mathcal{U}(\rho_{\min},\rho_{\max})$. Let $h\times w$ be the latent spatial size. We construct a square mask by
\begin{equation}
\begin{aligned}
\ell
&=
\min\!\left(
\min(h,w),
\max\!\left(1,
\operatorname{round}\!\left(\sqrt{\rho hw}\right)
\right)
\right),\\
a
&\sim\mathcal{U}\{0,\ldots,h-\ell\},
\qquad
b\sim\mathcal{U}\{0,\ldots,w-\ell\},\\
M[i,j]
&=
\mathbb{I}
\left[
a\leq i<a+\ell
\ \land\
b\leq j<b+\ell
\right].
\end{aligned}
\label{app:eq:square_mask}
\end{equation}
The mask is copied across latent channels to obtain $\bar M$. An entry of one marks a position selected for corruption and for evaluating the diffusion loss.

\subsection{Two-Stage Training}
\label{app:two_stage_training}

\renewcommand{\algorithmicrequire}{\textbf{Input:}}
\renewcommand{\algorithmicensure}{\textbf{Output:}}
\begin{algorithm*}[t]
\caption{Two-Stage Training of DMD}
\label{alg:dmd_training}
\footnotesize
\begin{algorithmic}[1]
\REQUIRE Normal dataset $\mathcal{D}_{N}$, epoch counts $E_1$ and $E_2$, diffusion horizon $T_d$, mask range $[\rho_{\min},\rho_{\max}]$, and loss weight $\lambda$
\ENSURE Trained $E_{\phi}$, $D_{\eta}$, $\mathcal{C}$, $\epsilon_{\omega}$, and $f_{\psi}$
\STATE Initialize $E_{\phi}$, $D_{\eta}$, and $\mathcal{C}$
\STATE \textbf{Stage I: quantized autoencoder pretraining}
\FOR{$e=1$ to $E_1$}
    \FOR{each batch $\{x_i\}_{i=1}^{B}\subset\mathcal{D}_{N}$}
        \STATE Encode, quantize each latent position, apply the straight-through estimator, and decode
        \STATE Compute $\mathcal{L}_{\mathrm{VQ}}$ and update $\phi$, $\eta$, and $\mathcal{C}$
    \ENDFOR
\ENDFOR
\STATE Fix $E_{\phi}$, $D_{\eta}$, and $\mathcal{C}$
\STATE Initialize $\epsilon_{\omega}$ and $f_{\psi}$
\STATE \textbf{Stage II: masked diffusion and reconstruction-shift discrimination}
\FOR{$e=1$ to $E_2$}
    \FOR{each batch $\{x_i\}_{i=1}^{B}\subset\mathcal{D}_{N}$}
        \STATE Compute $z_{0,i}=\operatorname{sg}[\mathcal{Q}_{\mathcal{C}}(E_{\phi}(x_i))]$
        \STATE Independently sample $M_i$, $t_i\sim\mathcal{U}\{1,\ldots,T_d\}$, and $\epsilon_i\sim\mathcal{N}(0,I)$ for every sample
        \STATE Apply selective noising inside $\bar M_i$ to obtain $z_{t_i}^{M_i}$
        \STATE Predict the noise and compute the one-step clean-latent estimate $\hat z_{0,i}^{M_i}$
        \STATE Decode $x_i^r=D_{\eta}(\hat z_{0,i}^{M_i})$
        \STATE Compute $\mathcal{L}_{\mathrm{diff}}$, $\mathcal{L}_{\mathrm{cls}}$, and $\mathcal{L}_{\mathrm{stage2}}=\mathcal{L}_{\mathrm{diff}}+\lambda\mathcal{L}_{\mathrm{cls}}$
        \STATE Update $\omega$ and $\psi$ only
    \ENDFOR
\ENDFOR
\STATE \textbf{return} $E_{\phi}$, $D_{\eta}$, $\mathcal{C}$, $\epsilon_{\omega}$, and $f_{\psi}$
\end{algorithmic}
\end{algorithm*}

During the second stage, $D_{\eta}$ is fixed but remains in the computation graph. The reconstructed counterpart $x^r$ is not detached. Therefore, $\mathcal{L}_{\mathrm{cls}}$ updates $\psi$ and propagates through $D_{\eta}$ to $\omega$, while $\phi$, $\eta$, and $\mathcal{C}$ remain unchanged. The classifier uses label zero for $x$ and label one for $x^r$, so $f_{\psi}$ denotes the reconstruction-shift probability.

\subsection{Dual-Level Inference}
\label{app:dual_level_inference}

All inference masks use side length $\ell_M$ and stride $s_M$, where $1\leq s_M\leq\ell_M$. Mask starts follow the regular grid and the final start along each dimension is set to $h-\ell_M$ or $w-\ell_M$. This boundary rule covers the complete latent grid. The resulting number of masks $K_M$ is determined by the grid rather than tuned independently.

\begin{algorithm*}[t]
\caption{Dual-Level Anomaly Inference}
\label{alg:dmd_inference}
\footnotesize
\begin{algorithmic}[1]
\REQUIRE Test image $x^*$, restoration step $t^*$, mask side length $\ell_M$, mask stride $s_M$, and constant $\delta$
\ENSURE Image-level score $S_{\mathrm{img}}(x^*)$ and pixel-level map $A_{\mathrm{pix}}(x^*)$
\STATE Compute $S_{\mathrm{img}}(x^*)=f_{\psi}(x^*)$
\STATE Compute $z_0^*=\mathcal{Q}_{\mathcal{C}}(E_{\phi}(x^*))$
\STATE Construct overlapping masks $\{M_k\}_{k=1}^{K_M}$ that cover the latent grid
\STATE Initialize $N_{\mathrm{pix}}=0_{H\times W}$ and $D_{\mathrm{pix}}=0_{H\times W}$
\FOR{$k=1$ to $K_M$}
    \STATE Sample $\epsilon_k\sim\mathcal{N}(0,I)$ and form the selectively noised latent $z_{t^*}^{M_k}$
    \STATE Use one denoiser evaluation to obtain $\hat z_0^{M_k}$ and decode $x_k^r=D_{\eta}(\hat z_0^{M_k})$
    \STATE Compute the channel-averaged residual $R_k$ and upsample $M_k$ by nearest-neighbor interpolation
    \STATE Update $N_{\mathrm{pix}}\leftarrow N_{\mathrm{pix}}+M_k^{\uparrow}\odot R_k$
    \STATE Update $D_{\mathrm{pix}}\leftarrow D_{\mathrm{pix}}+M_k^{\uparrow}$
\ENDFOR
\STATE Compute $A_{\mathrm{pix}}(x^*)=N_{\mathrm{pix}}/(D_{\mathrm{pix}}+\delta)$
\STATE \textbf{return} $S_{\mathrm{img}}(x^*)$ and $A_{\mathrm{pix}}(x^*)$
\end{algorithmic}
\end{algorithm*}

The restoration step $t^*$ is the noise level used by the one-step clean-latent estimator, not the number of reverse iterations. We use one Gaussian draw per mask and fix the inference seed for the reported evaluation.

\section{Details of medical anomaly detection datasets}
\label{app:datasets}

\paragraph{Dataset rationale.}
Our dataset selection follows two principles: comparability with established medical anomaly detection benchmarks and coverage of heterogeneous imaging modalities. For brain MRI, we use the BraTS2021-derived split from MedIAnomaly~\cite{cai2025medianomaly}, because it provides a standardized normal-only training and mixed normal/abnormal testing protocol that has been used to benchmark multiple reconstruction- and diffusion-based anomaly detection methods. Although newer BraTS releases, such as BraTS 2024/2025, contain larger and more diverse challenge data, they are organized around segmentation-oriented clinical tasks, including pre/post-treatment glioma, metastases, meningioma, pediatric tumors, missing data, generalizability, MRI synthesis, and inpainting, rather than the same unsupervised anomaly detection protocol. Therefore, we use the BraTS2021-based MedIAnomaly split to ensure fair comparison with prior anomaly detection baselines instead of mixing incompatible dataset protocols.

Table~\ref{tab:dataset_summary} summarizes the dataset statistics, and we briefly introduce each dataset below:
\begin{itemize}[leftmargin=*]
\item \textbf{BraTS2021:} The original BraTS2021 dataset contains MRI cases with voxel-level tumor annotations~\cite{RN21}. Following the MedIAnomaly benchmark protocol~\cite{cai2025medianomaly}, we use the BraTS2021-derived anomaly detection split, which contains 4,211 tumor-free images for training and 2,776 images for testing, including 828 normal samples and 1,948 abnormal. This split is used for comparability with prior unsupervised medical anomaly detection methods.

\item \textbf{IXI/ATLAS 2.0:} The training set combines 582 T1-weighted MRI scans from IXI~\cite{IXI2023} and 217 healthy samples from ATLAS v2.0~\cite{liew2022large}. For evaluation, we use 655 ATLAS T1-weighted scans with expert-annotated lesion masks.

\item \textbf{WMH:} The WMH dataset contains FLAIR and T1 scans from the White Matter Hyperintensity segmentation challenge~\cite{kuijf2022data}. We use 110 2D multi-slice FLAIR scans for cross-dataset evaluation with the BraTS2021-trained model, testing robustness under domain shift.

\item \textbf{BUSI:} BUSI~\cite{al2020dataset} contains breast ultrasound images from female aged 25--75 years. Following the low-data setting used in our protocol, 32 normal images are used for training and 647 images are used for testing, including 437 benign and 210 malignant lesion images. Because the training set is small, we report averaged results over multiple runs and discuss this setting as a low-data stress test rather than the sole evidence for generalization.

\item \textbf{VinDr-CXR:} VinDr-CXR~\cite{nguyen2022vindr} is a large-scale chest radiograph dataset with normal and abnormal X-ray images. Following MedIAnomaly~\cite{cai2025medianomaly}, we use 4,000 normal images for training and 2,000 images for testing, including 1,000 normal and 1,000 abnormal samples.
\end{itemize}

\begin{table}[ht]
\setlength{\tabcolsep}{1mm}
\centering
\setlength{\tabcolsep}{0.6mm}
\begin{tabular}{lcccc}
\toprule
\textbf{Dataset} & \textbf{Modality} & \textbf{Train}  & \textbf{Test} &\textbf{Size}\\
\midrule
BraTS2021 & FLAIR& 4,211  & 2776 &240\\
IXI/ATLAS & T1-weighted  & 582 + 217 & 655  &256/182   \\
WMH  & FLAIR & 0 & 110 &240\\
BUSI  &  Ultrasound & 32 & 647 & 512  \\
VinDr-CXR  & X-ray & 4,000 & 2,000 & 256 \\
\bottomrule
\end{tabular}
\caption{Summary of main characteristics of each dataset used in the experiment.}
\label{tab:dataset_summary}
\end{table}

\section{More implementation details}
\label{app:implementation_details}
\paragraph{Details of Evaluation Metrics}
\label{app:metrics}
For image-level anomaly detection, AUC evaluates the ranking of normal and abnormal images across all decision thresholds, while AP summarizes the precision--recall curve and is particularly informative under class imbalance. Both metrics are computed directly from continuous anomaly scores. F1 is computed using a fixed operating threshold set to the 95th percentile of the anomaly scores obtained from normal training images. For pixel-level anomaly localization, AP$_{\mathrm{pix}}$ evaluates the ranking quality of continuous pixel-wise anomaly scores. Dice measures the overlap between the predicted binary anomaly map and the ground-truth lesion mask. The binarization threshold is set to the 99.5th percentile of pixel-level anomaly scores obtained from normal training images. The same threshold-calibration rules are applied to all datasets and runs, without using test labels or lesion annotations.

For brain imaging datasets, we additionally adopt the normative
reconstruction metrics proposed by Bercea et al.~\cite{bercea2025evaluating}. The Restoration Quality Index (RQI) evaluates the perceptual fidelity of reconstructions for normal images. The Anomaly-to-Healthy Index (AHI) measures how closely reconstructed pathological images approach a healthy reference distribution. The Healthy Conservation and Anomaly Correction Index (CACI) evaluates the ability to preserve healthy regions while correcting abnormal regions.

\paragraph{VQ-VAE Architecture.}
The VQ-VAE encoder uses convolutional blocks with channel progression $1\rightarrow32\rightarrow64\rightarrow128$. Each block applies a strided convolution for spatial downsampling, followed by BatchNorm and ReLU. The decoder follows a symmetric structure and uses transposed convolutions for upsampling. The codebook contains $K=256$ entries, and each code embedding has dimension $d_z=64$. After Stage I, the encoder, decoder, and codebook are fixed.

\paragraph{Latent Diffusion Module.}
The diffusion model operates on the quantized code embeddings produced by the VQ-VAE. We use an attention-enhanced denoising U-Net as the noise prediction network. The diffusion horizon is $T_d=1000$, with a linear $\beta_t$ schedule from $1\times10^{-4}$ to $0.02$. During Stage II, one square mask is sampled for each latent representation. Gaussian noise is added only at the selected latent positions, while the remaining positions provide a clean context. The denoiser predicts the injected noise from $z_t^M$ and $t$. The diffusion loss is averaged over the selected positions and latent channels. A single denoiser evaluation produces the clean-latent estimate used by the fixed decoder.

\paragraph{Reconstruction-Shift Classifier.}
The classifier contains three convolutional layers with 32, 64, and 128 output channels. Each convolutional layer is followed by BatchNorm, ReLU, and pooling. The resulting features are passed through two fully connected layers and a two-class softmax. We use the class-one probability as $f_{\psi}$. Original normal images have label zero, and their reconstructed counterparts have label one.

\paragraph{Parameter Selection.}
The training mask range $[\rho_{\min},\rho_{\max}]$, inference parameters $\ell_M$, $s_M$, and $t^*$, and decision thresholds are selected on the validation split and then fixed for testing. No test label or test mask is used for parameter selection.

\section{Evaluation of Normative Reconstruction Quality}
\label{app:normative_quality}

We further evaluate the normative reconstruction quality of different methods using three complementary metrics, namely RQI, AHI, and CACI, on BraTS2021, ATLAS 2.0, and WMH. As shown in Table~\ref{tab:normative_multi}, DMD achieves consistently strong performance across the three brain imaging benchmarks. On BraTS2021, DMD obtains the best results on all three metrics, with RQI, AHI, and CACI scores of 0.88, 0.53, and 0.62, respectively. Compared with AutoDDPM, DMD improves RQI by 0.02, AHI by 0.05, and CACI by 0.15, suggesting better reconstruction fidelity, stronger healthy-distribution alignment, and improved correction-preservation balance. On ATLAS 2.0, AutoDDPM achieves a slightly higher RQI score, while DMD obtains the best AHI and CACI scores. This indicates that although AutoDDPM provides marginally higher reconstruction fidelity, DMD better normalizes abnormal regions toward the healthy distribution and better preserves healthy structures during correction. On WMH, DMD achieves the highest RQI and CACI scores, while AutoDDPM obtains the highest AHI. These results suggest that DMD maintains robust normative reconstruction quality under cross-dataset domain shift, particularly in terms of preserving normal tissue while correcting abnormal regions.

\begin{table*}[t]
\centering
\setlength{\tabcolsep}{2.2mm}
\resizebox{\textwidth}{!}{%
\begin{tabular}{lccccccccc}
\toprule
\multirow{2}{*}{\textbf{Method}}
& \multicolumn{3}{c}{\textbf{BraTS2021}}
& \multicolumn{3}{c}{\textbf{ATLAS 2.0}}
& \multicolumn{3}{c}{\textbf{WMH}} \\
\cmidrule(r){2-4} \cmidrule(r){5-7} \cmidrule(r){8-10}
& RQI$\uparrow$ & AHI$\uparrow$ & CACI$\uparrow$
& RQI$\uparrow$ & AHI$\uparrow$ & CACI$\uparrow$
& RQI$\uparrow$ & AHI$\uparrow$ & CACI$\uparrow$ \\
\midrule
VAE
& 0.04 & 0.00 & 0.19
& 0.06 & 0.00 & 0.06
& 0.12 & 0.00 & 0.08 \\
f-AnoGAN
& 0.62 & 0.00 & 0.32
& 0.43 & 0.00 & 0.23
& 0.56 & 0.00 & 0.18 \\
AutoDDPM
& 0.86 & 0.48 & 0.47
& \textbf{0.87} & 0.43 & 0.45
& 0.75 & \textbf{0.46} & 0.41 \\
\midrule
\textbf{DMD}
& \textbf{0.88} & \textbf{0.53} & \textbf{0.62}
& 0.84 & \textbf{0.48} & \textbf{0.52}
& \textbf{0.79} & 0.44 & \textbf{0.48} \\
\bottomrule
\end{tabular}
}
\caption{Additional evaluation of normative reconstruction quality on BraTS2021, ATLAS 2.0, and WMH using RQI, AHI, and CACI. The best results are marked in \textbf{bold}.}
\label{tab:normative_multi}
\end{table*}

\begin{figure}[!ht]
    \centering
    \includegraphics[width=\linewidth]{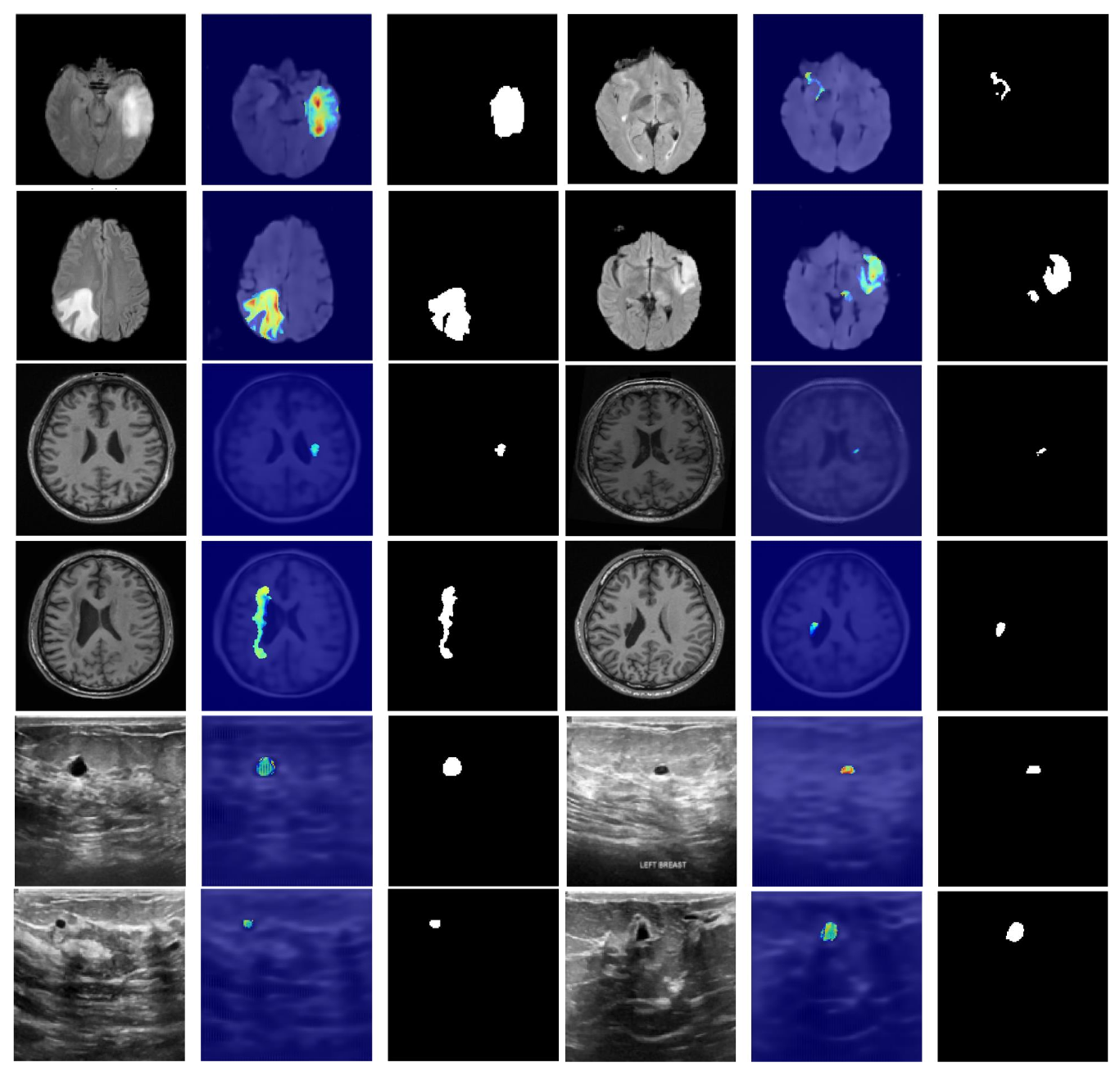}
    \caption{More visual comparison of anomaly detection results on BraTS2021 (row 1-2), ATLAS2.0 (row 3-4) and BUSI (row 5-6) datasets using our proposed approach. The first and fourth column shows the original input images, the second and fifth column is the anomaly map, and the third and sixth column is the ground truth.}
    \label{fig:visual supp}
\end{figure}

\section{Additional Visualizations Result}
\label{app:additional_visualizations}
To further validate the effectiveness and generalization of our method, we provide additional qualitative results on the BraTS2021, ATLAS2.0 and BUSI datasets in Figure~\ref{fig:visual supp}. As shown, each row presents a representative case: the first and fourth column displays the input images, the second and fifth column shows the anomaly map generated by our approach, and the third and sixth column provides the ground truth mask. The qualitative results demonstrate that our method produces accurate and consistent anomaly detection, effectively reconstructing healthy regions while accurately highlighting abnormal areas. By comparing the predicted anomaly maps with the ground-truth masks, we can clearly observe the effectiveness of our method, which is capable of accurately recovering almost all anomalous regions delineated by the masks.

\section{Additional Parameter Sensitivity Analysis}
\label{app:parameter_sensitivity}
We further analyze the sensitivity of DMD to two key hyperparameters: the number of diffusion timesteps $t$ and the mask region size. Each configuration is repeated three times with different random seeds, and the mean Dice score is reported with the corresponding standard deviation.

\begin{figure}[t]
    \centering
    \begin{subfigure}[t]{0.45\linewidth}
        \centering
        \includegraphics[width=\linewidth, height=0.12\textheight]{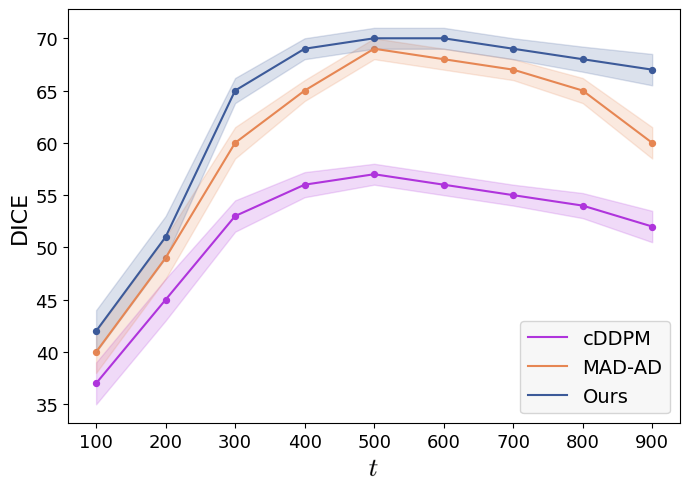}
        \caption{Diffusion timesteps $t$.}
        \label{fig:param-timestep}
    \end{subfigure}
    \hfill
    \begin{subfigure}[t]{0.45\linewidth}
        \centering
        \includegraphics[width=\linewidth, height=0.12\textheight]{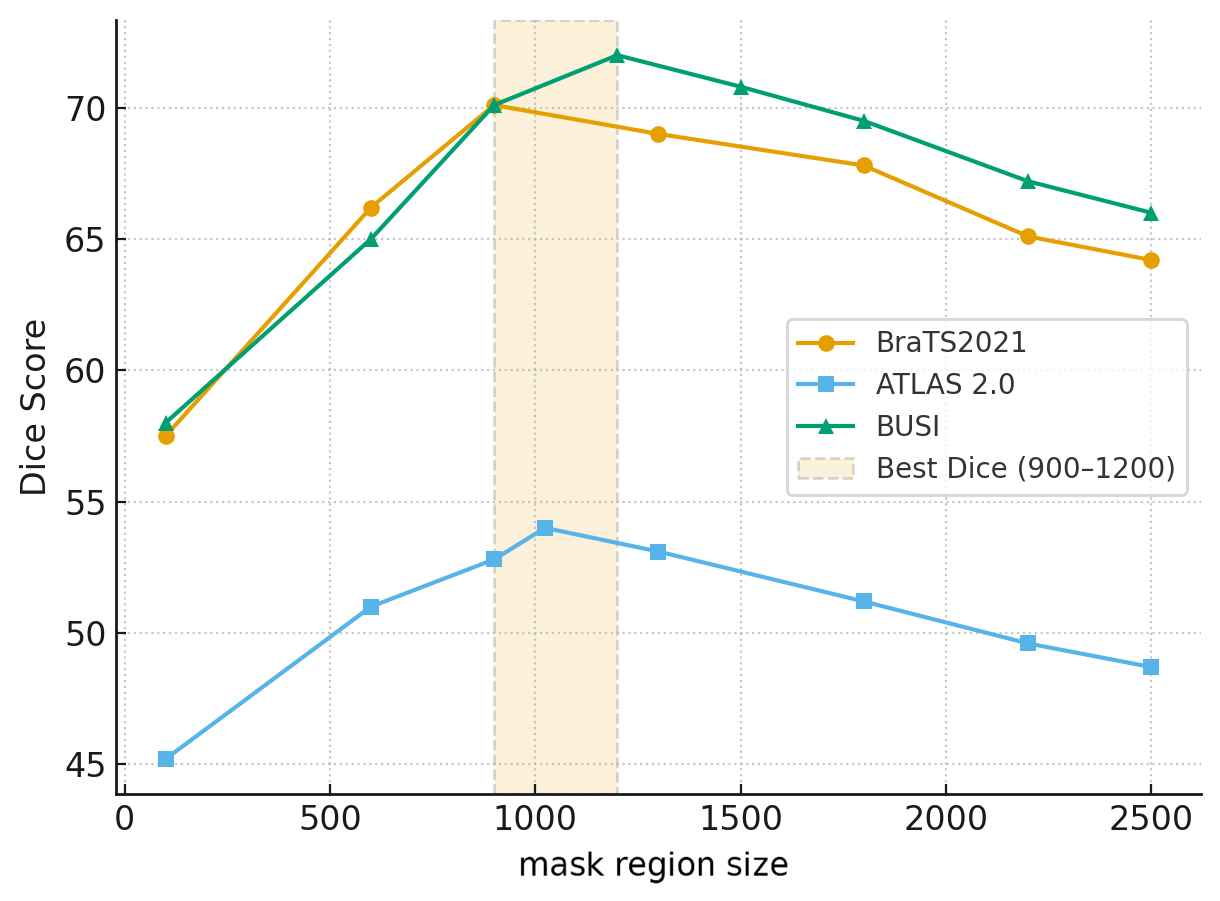}
        \caption{Mask region size.}
        \label{fig:param-mask}
    \end{subfigure}
    \caption{Parameter sensitivity analysis of DMD.
(a) Dice scores on BraTS2021 under different diffusion timesteps $t$, compared with cDDPM and MAD-AD.
(b) Dice scores under different mask region sizes on BraTS2021, ATLAS 2.0, and BUSI. The shaded region indicates the empirically stable range of mask sizes.}
    \label{fig:param-sensitivity}
\end{figure}

\begin{figure}[!t]
    \centering
    \includegraphics[width=0.8\linewidth]{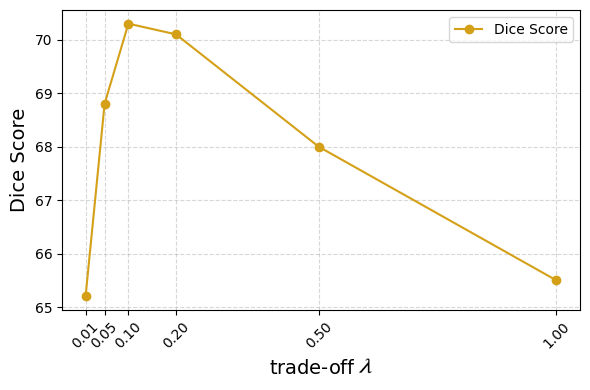}
    \caption{Trade-off $\lambda$ Parameter sensitivity analysis on BraTS2021}
    \label{fig:param-beta}
\end{figure}
\paragraph{Effect of Diffusion Timesteps and Mask Region Size}
Figure~\ref{fig:param-timestep} shows the effect of diffusion timesteps on BraTS2021. The Dice score increases as $t$ grows from 100 to 500, suggesting that moderate denoising improves pseudo-healthy reconstruction and anomaly localization. However, performance decreases when $t$ becomes too large, indicating that excessive diffusion steps may introduce unnecessary perturbations or over-smooth localized abnormal structures. Across all timestep settings, DMD consistently outperforms cDDPM and MAD-AD, with the best performance observed around $t=500$--$600$. Figure~\ref{fig:param-mask} evaluates the effect of mask region size on BraTS2021, ATLAS 2.0, and BUSI. Across datasets, performance generally improves when the mask size increases from small regions to a moderate range, indicating that localized latent masking provides useful perturbations for learning pseudo-healthy reconstruction. The best overall performance is achieved when the mask region size is approximately $900$--$1200$ pixels. When the mask becomes too large, the Dice score declines, suggesting that overly aggressive masking may remove excessive anatomical context and weaken the effectiveness of localized guidance.

\paragraph{Effect of the Discriminative Loss Weight}
To assess the effect of the trade-off parameter $\lambda$ on model performance, we conducted a sensitivity analysis by varying $\lambda$ while holding all other hyperparameters fixed. As shown in Figure~\ref{fig:param-beta}, the Dice score initially increases as $\lambda$ grows, reflecting the positive impact of including classification loss. However, when $\lambda$ becomes excessively large, segmentation performance begins to degrade, indicating that overemphasis on the discriminative objective undermines reconstruction quality. The best performance is observed at $\lambda = 0.1$, where the balance between generative and discriminative learning is optimal.

\end{document}